\documentclass{article}

\usepackage[dblblindworkshop, final]{neurips_2026}
\usepackage{multirow}
\usepackage{booktabs}
\usepackage{xcolor}
\usepackage{amssymb}
\usepackage{amsmath}

\workshoptitle{Linguistic Principles for Foundation Models}

\usepackage[utf8]{inputenc} 
\usepackage[T1]{fontenc}    
\usepackage{hyperref}       
\usepackage{url}            
\usepackage{booktabs}       
\usepackage{amsfonts}       
\usepackage{nicefrac}       
\usepackage{microtype}      
\usepackage{xcolor}         
\usepackage{graphicx}

\title{Better Boundary Prediction Does Not Guarantee Better Structural Representation}

\author{%
  Michael Zhou\textsuperscript{1},
  Markus Frohmann\textsuperscript{2}, Junghyun Min\textsuperscript{3}, En-Shiun Annie Lee\textsuperscript{2,4}\\
  \textsuperscript{1}Carnegie Mellon University, \textsuperscript{2}University of Toronto \\
  \textsuperscript{3}Georgetown University,
  \textsuperscript{4}Ontario Tech University\\
  \texttt{mjzhou@andrew.cmu.edu} \\
}

\begin{document}

\maketitle

\begin{abstract}
Punctuation marks often mark phrase boundaries, carrying linguistic information.
Therefore, punctuation marks' abundance and feasibility as proxies of linguistic structure has led to suggestions that using punctuation marks to optimise on phrasal structure may help language models (LMs) improve their \textit{representations of argument, semantic, and syntactic structures}.
In this paper, we test the above claim in English across three primary LM architectures: decoder-only, encoder-only, and encoder-decoder LMs, by complementing traditional token prediction with \textbf{punctuation restoration (PR)}.
Our evidence, collected at the $<1$B parameter scale, suggests that optimising for boundary prediction does not necessarily guarantee higher-quality structural representation in decoder-only, encoder-decoder, and encoder-only LMs, failing to improve performance on downstream tasks that require understanding structure in linguistic input.
\end{abstract}

\section{Introduction}
While modern language models (LMs) already perform well in most NLP tasks, there is room for improvement in understanding structure in text, as LMs trained on token prediction often rely on linear surface patterns rather than the underlying structures \citep{mccoy-etal-2019-right, dziri2023faith} and do not fully capture the hierarchical structure in linguistic input, leading to failures in retrieval \citep[e.g.][]{berglund2024reversal, kitouni2024factorization} and reasoning \citep[e.g.][]{levine-zeldes-2026-cracks}\footnote{See \citet{celsi2026beyond} for a comprehensive overview of autoregressive LLM failure modes.}.

Punctuation often encodes vital structural boundaries in discourse, syntax, and meaning \citep{dale1991exploring, briscoe1996syntax}, carrying structural information often attributed to prosodic features in speech \citep{yadavalli-etal-2026-prosody, min2026usingprosodypredictsyntactic}.
Punctuation marks thus have been proposed to serve as implicit structural cues in LM pre-training: \citet{min-etal-2025-punctuation} have demonstrated that additional representation learning via continual pre-training on punctuation restoration (PR) improves language models' performance in structure-related NLP tasks.

However, the mechanisms of using PR as a learning objective are not fully understood, as previous work only investigates a single transformer architecture \citep[T5;][]{raffel2023exploringlimitstransferlearning} using proprietary training data.
There may be unintended effects of extended pre-training, such as overtraining \citep{springer2025overtrainedlanguagemodelsharder} that may lead to degradation in future task-specific fine-tuning, and model misalignment \citep{betley2026training}.
In this paper, we perform a wide-scale, systematic evaluation into the effects of additional pre-training on PR for structure learning and natural language understanding.
Working with English, we investigate the relationship between boundary prediction and structural representation quality, as measured by downstream task performance, across three architectures and a wider range of evaluation tasks.
As hypothesized in previous work and supported by the empirical connection between punctuation marks and linguistic structure, we predict that optimisation on boundary prediction (with PR as its proxy) will result in higher quality structural representation, measured by downstream task performance.

Contrary to our prediction, our empirical results across 3 transformer architectures and 17 tasks suggests that simply optimising for better boundary prediction does not necessarily result in better structure understanding; additional pre-training on PR does \textbf{not} reliably result in greater downstream task performance despite sizeable improvement in PR performance.
We interpret our results to indicate that PR is an insufficient proxy for inducing structural understanding.
While models can successfully learn the local, surface-level patterns required to predict missing punctuation, this shallow boundary detection does not automatically translate into the rich, hierarchical representations necessary for natural language understanding.\footnote{We release code and weights at \url{https://github.com/Michael-Z-J/pr_gpt2_bert} to facilitate future work.}


\section{Methods and Experimental Setup}
\subsection{Using Punctuation Restoration as a Training Objective}
\label{sec:pr-train-obj}
We test the PR learning objective across three transformer neural network architectures. The decoder-only architecture is GPT-2 \citep{radford2019language}, the encoder-only model BERT \citep{devlin2019bertpretrainingdeepbidirectional}, and the encoder-decoder model T5 \citep{raffel2023exploringlimitstransferlearning}. All models are trained on approximately 300k financial news articles sourced from five US-based outlets from January to May of 2018\footnote{Accessed via \url{https://huggingface.co/datasets/ashraq/financial-news-articles}.}. We train the model to restore eight punctuation marks: period, comma, exclamation mark, question mark, semicolon, colon, quotation mark and apostrophe. We selected these punctuation marks due to their usage frequency \citep{sun2019frequency}, their role in semantic structure, rich information about discourse relations, and grammatical mood \citep{dale1991exploring, briscoe1996syntax}.

We follow \citet{min-etal-2025-punctuation} in constructing our training data.
For each example, the original text is stripped of capitalisation and selected punctuation marks. 
To make the examples, we use a sliding window on each article to ensure that we efficiently use the entirety of the text. For each window, we create an example before shifting the window along the article, where adjacent windows overlap. There are several architecture-specific implementation differences that we describe below.

\paragraph{Encoder-decoder T5-base.} PR is implemented as a sequence-to-sequence task for T5. The model is given examples stripped of the selected punctuation marks and capitalisation as input, and the target sequence is the original example.
We find that gradient accumulation is an essential hyperparameter for convergence, due to the sparsity of punctuation marks (around 1 in 20 tokens) in our corpus.
To evaluate PR, generated words are aligned with the target sequence, to determine if the model associated the correct punctuation and capitalisation to each word.

\paragraph{Decoder-only GPT-2-medium.} We treat PR as a conditional text generation task for GPT-2. GPT-2 receives the prompt "restore punctuation: <source> restored: ", where <source> is an example with capitalisation and selected punctuation removed. Each of the tokens in the prompt is assigned labels of $-$100, excluding it from loss computation and evaluation.
The prompt is only used to provide context. Evaluation of the generated text is done in the same manner as for T5's text generation, and we record the F1 score for PR.

\paragraph{Encoder-only BERT-base-uncased.} We implement PR as a classification task with two classification heads for BERT.
The first head produces a multiclass classification regarding capitalisation.
The label classes are whether the word is: all uppercase, all lowercase, uppercase-initial, or mixed (e.g. PhD). 
The second head consists of eight binary classifiers, corresponding to the prediction on each selected punctuation mark. 
Each word is associated with a set of binary punctuation targets and a multiclass capitalisation target. We train the model to minimise the combined loss of the capitalisation classification and the punctuation classification. F1 of the PR performance is measured as the average between F1 of capitalisation head and F1 of punctuation head.

We evaluated all models periodically to observe progress. We also used early stopping with a patience of three evaluations to reduce overfitting and computation waste. For the other settings, we used model-specific values for parameters such as learning rate and maximum sequence length so that training is appropriate for each model.

\subsection{Downstream Task Evaluation}
\label{sec:downstream-task-eval}
The primary objective is to analyse the impact of the learning objective on linguistic structure understanding.
For each of the three architectures (BERT, GPT-2, T5), we compare off-the-shelf checkpoints (base) against the continually pre-trained counterpart using PR (+PR). 
To evaluate linguistic structure understanding, we use a range of NLP tasks, covering structure-related tasks (named entity recognition, relation extraction, part-of-speech tagging, semantic role labeling, open information extraction, largely following \citet{lee2025structured}), and general language understanding evalaution tasks \citep[GLUE;][]{wang-etal-2018-glue}, to analyse the effects of PR. The primarily hypothesized benefits gained by extended pretraining on PR is learning the structural and linguistic information in text. The tasks that were selected test the model's ability to identify entities in the text, determine relations between entities, among other understanding tasks.

We run each experiment three times with different seed to test statistical significance of any difference in performance. We use each dataset's default splits. We choose commonly-used hyperparameters for each architecture (e.g. learning rate for GPT-2: $5\times10^{-5}$, BERT: $5\times10^{-5}$, T5: $3\times10^{-4}$). To obtain comparable results, hyperparameters for base and +PR models are kept the same.

\paragraph{Structure-related tasks.}
We include five structure-related tasks as part of our evaluation, namely relation extraction (RE), named entity recognition (NER), part-of-speech tagging (POS), semantic role labeling (SRL), and open information extraction (OpenIE).

\textbf{RE} is a classification task.
Given a sentence and a pair of entities within the sentence, the goal is to detect and accurately label the relationship between the two entities.
We use TACRED \citep{tacred} as our RE dataset. 
\textbf{NER} is a task of identifying entities within a text. Possible labels include person, location, and organisation.
To evaluate NER performance of the models, we use GENIA \citep{kim2003genia}, CoNLL 2003 \citep{tjong-kim-sang-de-meulder-2003-introduction}, and OntoNotes5 \citep{hovy-etal-2006-ontonotes}. 
In the \textbf{POS} task, the model maps each word to its part of speech. For POS, we use CoNLL 2000 \citep{sang2000introduction} with a total of 44 labels, based on the Penn Treebank tags. 
\textbf{SRL} is a task of identifying the components of a predicate-argument structure. 
Given a predicate, the model identifies the semantic roles of the surrounding nouns and assigns semantic role labels.
We use CoNLL 2012 \citep{pradhan-etal-2012-conll}.
\textbf{OpenIE} is a task of extracting structured relational triples (subject, relation, object) from unstructured natural language text.
Unlike in RE, the list of relations is not pre-defined.
We use OIE2016 \citep{stanovsky-dagan-2016-creating}.

\paragraph{General language understanding.}
To measure a model's general language understanding, we use the GLUE tasks \citep{wang-etal-2018-glue}.\footnote{We follow previous work in using the GLUE dataset's validation split for evaluation.} 
The GLUE benchmark tasks consist of a collection of single-sentence tasks, similarity and paraphrase tasks, and inference tasks.

The single-sentence tasks are CoLA \citep{warstadt2018neural} and SST-2 \citep{socher2013recursive}. For CoLA, when given a sentence, the model must determine whether it is linguistically acceptable or not. For SST-2, the model must determine whether a sentence has positive or negative sentiment.
The similarity and paraphrase tasks are MRPC \citep{dolan2005automatically}, STS-B \citep{cer2017semeval} and QQP\footnote{\url{ https://data.quora.com/First-Quora-Dataset-Release-Question-Pairs}}.
MRPC and QQP require the model to determine if two sentences are paraphrases of each other. 
For STS-B, given two sentences, the model returns similarity between the sentences on a scale from 0 to 5.
Finally, the inference tasks are MNLI \citep{williams-etal-2018-broad}, QNLI \citep{rajpurkar-etal-2016-squad}, RTE \citep{dagan2006pascal},
and WNLI \citep{levesque2011winograd}. These tasks require the model, when given a premise and a hypothesis, to determine if there is entailment. 

\subsection{FLOPs and Data Control}
Our research question asks whether punctuation restoration as a learning objective affects structure representation in small language models.
To investigate this, we continually train off-the-shelf language models using punctuation restoration as we describe in Section \ref{sec:pr-train-obj}.
However, it is possible that any effects from this be attributable to additional training---to either exposure to novel data or additional floating point operations (FLOPs)---rather than using PR as a training objective.
Thus, we implement a continually trained control on language modeling to separate the effects of additional training in floating point operations from that of punctuation restoration.
Using our training corpus of financial news articles, we continually pre-train T5 using the language modeling objective.



\section{Results and Discussion}
\label{sec:results-discussion}

\begin{table*}[!t]
\centering
\small

\begin{tabular}{l ccl ccl ccl}
\toprule
& \multicolumn{3}{c}{\textbf{T5-base}}
& \multicolumn{3}{c}{\textbf{GPT-2-medium}}
& \multicolumn{3}{c}{\textbf{BERT-base}} \\
\cmidrule(lr){2-4}
\cmidrule(lr){5-7}
\cmidrule(lr){8-10}

\textbf{Task (Dataset)}
& \textbf{base} & \textbf{+PR} & \multicolumn{1}{c}{$\Delta$}
& \textbf{base} & \textbf{+PR} & \multicolumn{1}{c}{$\Delta$}
& \textbf{base} & \textbf{+PR} & \multicolumn{1}{c}{$\Delta$} \\
\midrule


PR
& 74.0 & 85.0 & {\color{green}$\blacktriangle$}11.0
& 74.1 & 90.0 & {\color{green}$\blacktriangle$}15.9
& 79.0 & 89.0 & {\color{green}$\blacktriangle$}10.0 \\

\midrule
\multicolumn{10}{l}{\textbf{Structure-related tasks}} \\

RE (TACRED)
& 63.1 & 60.5 & {\color{red}$\blacktriangledown$}2.67$^{*}$
& 64.3 & 65.6 & {\color{green}$\blacktriangle$}1.28$^{*}$
& 66.0 & 64.5 & {\color{red}$\blacktriangledown$}1.50 \\

NER (CoNLL 2003)
& 89.4 & 89.1 & {\color{red}$\blacktriangledown$}0.31
& 83.0 & 78.7 & {\color{red}$\blacktriangledown$}4.27
& 90.9 & 91.0 & {\color{green}$\blacktriangle$}0.19 \\

NER (GENIA)
& 77.7 & 77.3 & {\color{red}$\blacktriangledown$}0.40
& 48.3 & 49.1 & {\color{green}$\blacktriangle$}0.77
& 76.4 & 75.3 & {\color{red}$\blacktriangledown$}1.12$^{*}$ \\

NER (OntoNotes5)
& 90.5 & 90.1 & {\color{red}$\blacktriangledown$}0.41
& 79.0 & 78.3 & {\color{red}$\blacktriangledown$}0.77$^{*}$
& 86.9 & 86.8 & {\color{red}$\blacktriangledown$}0.15 \\

POS (CoNLL 2000)
& 97.7 & 97.7 & {\color{red}$\blacktriangledown$}0.02
& 96.4 & 96.3 & {\color{red}$\blacktriangledown$}0.13
& 98.1 & 98.1 & {\color{green}$\blacktriangle$}0.03 \\

SRL (CoNLL 2012)
& 87.5 & 87.1 & {\color{red}$\blacktriangledown$}0.36$^{*}$
& 82.3 & 82.2 & {\color{red}$\blacktriangledown$}0.10
& 88.3 & 87.9 & {\color{red}$\blacktriangledown$}0.31$^{*}$ \\

OpenIE(OIE2016)
& 66.7 & 64.9 & {\color{red}$\blacktriangledown$}1.85
& 47.3 & 45.1 & {\color{red}$\blacktriangledown$}2.20
& 84.2 & 82.8 & {\color{red}$\blacktriangledown$}1.38 \\

\midrule
\multicolumn{10}{l}{\textbf{General language understanding tasks}} \\
SST-2
& 93.2 & 92.0 & {\color{red}$\blacktriangledown$}1.19$^{*}$
& 93.9 & 92.7 & {\color{red}$\blacktriangledown$}1.26$^{*}$
& 92.5 & 90.6 & {\color{red}$\blacktriangledown$}1.95$^{*}$ \\

QQP
& 90.4 & 89.2 & {\color{red}$\blacktriangledown$}1.16$^{*}$
& 90.3 & 90.2 & {\color{red}$\blacktriangledown$}0.12
& 91.1 & 90.7 & {\color{red}$\blacktriangledown$}0.38$^{*}$ \\

MNLI-m
& 85.9 & 83.6 & {\color{red}$\blacktriangledown$}2.29$^{*}$
& 85.3 & 84.0 & {\color{red}$\blacktriangledown$}1.30$^{*}$
& 84.1 & 83.3 & {\color{red}$\blacktriangledown$}0.73 \\

MNLI-mm
& 86.2 & 84.4 & {\color{red}$\blacktriangledown$}1.81$^{*}$
& 85.4 & 84.5 & {\color{red}$\blacktriangledown$}0.92$^{*}$
& 84.6 & 83.6 & {\color{red}$\blacktriangledown$}1.00$^{*}$ \\

QNLI
& 90.7 & 88.5 & {\color{red}$\blacktriangledown$}2.17$^{*}$
& 90.3 & 89.6 & {\color{red}$\blacktriangledown$}0.66
& 91.1 & 89.6 & {\color{red}$\blacktriangledown$}1.48 \\

RTE
& 55.6 & 58.4 & {\color{green}$\blacktriangle$}2.77
& 66.5 & 59.4 & {\color{red}$\blacktriangledown$}7.10$^{*}$
& 64.6 & 61.0 & {\color{red}$\blacktriangledown$}3.61$^{*}$ \\

WNLI
& 49.8 & 52.1 & {\color{green}$\blacktriangle$}2.35
& 37.1 & 29.1 & {\color{red}$\blacktriangledown$}7.98
& 33.8 & 40.8 & {\color{green}$\blacktriangle$}7.04 \\

CoLA
& 44.1 & 36.0 & {\color{red}$\blacktriangledown$}8.12$^{*}$
& 49.7 & 43.6 & {\color{red}$\blacktriangledown$}6.06
& 58.2 & 51.8 & {\color{red}$\blacktriangledown$}6.39$^{*}$ \\

MRPC
& 70.8 & 73.7 & {\color{green}$\blacktriangle$}2.86$^{*}$
& 79.0 & 77.9 & {\color{red}$\blacktriangledown$}1.14
& 86.2 & 79.7 & {\color{red}$\blacktriangledown$}6.54$^{*}$ \\

STS-B
& 73.3 & 78.6 & {\color{green}$\blacktriangle$}5.23
& 85.2 & 82.7 & {\color{red}$\blacktriangledown$}2.44$^{*}$
& 89.1 & 87.1 & {\color{red}$\blacktriangledown$}1.97$^{*}$ \\

\bottomrule
\end{tabular}

\caption{
Punctuation restoration and downstream task performance.
PR ``base'' performance is measured after 1 epoch of training on PR.
Structure-related task performance is F1; CoLA, Matthews correlation; STS-B, Pearson correlation; all other GLUE tasks, accuracy.
We report the average across 3 runs with random classifier initialisation.
Using Welch's t-test, statistical significance $\Delta$ ($p<0.05$) is noted with an *.
}
\label{tab:downstream_and_glue_results}
\end{table*}

\paragraph{FLOPs and data control.}
Our FLOPs and data control overall degrades in downstream task performance, compared to the off-the-shelf version.
This suggests that the amount of FLOPs or the exposure to novel data is not the main driver for any potential performance gain that we may observe, and that any gain can be attributed to the training objective itself.
We detail the performance of this baseline in Appendix \ref{sec:appendix-flops-baseline}.

\paragraph{Effects of PR as a training objective.}
As expected, continual pre-training on punctuation restoration (PR) yields substantial improvements in boundary prediction capabilities. As shown in Table \ref{tab:downstream_and_glue_results}, PR macro F1 scores increased by $11\%$p for T5, $15.9\%$p for GPT-2, and $10\%$p for BERT. 

However, contrary to our initial hypothesis, this improvement in boundary prediction does not yield improved structure or general understanding.
Optimisation on PR does not reliably result in improved downstream task performance; in fact, across 3 transformer architectures and 17 tasks, we only observe 9 improvements out of 51 model-task combinations, with only 2 being statistically significant.
Performance on GLUE generally degraded, with the largest drops on CoLA ($-8.12$ for T5, $-6.39$ for BERT), RTE ($-7.10$ for GPT-2) and MRPC ($-6.54$ for BERT). Even on structure-focused tasks such as SRL and OpenIE, performance consistently declined across all three architectures.

We interpret this to indicate our titular message---that better boundary prediction does not guarantee better structural representation.
While we hypothesized that optimisation on surface-level features may activate deeper syntactic and semantic structures, the evidence suggests otherwise: transformer models simply learn to optimise on surface-level boundary prediction, rather than the deeper underlying structure in language.
This finding joins a suite of previous work documenting small language models' optimisation on shallow heuristics like lexical frequency and lexical overlap \citep{ gururangan-etal-2018-annotation, niven-kao-2019-probing, mccoy-etal-2019-right, zhou-etal-2020-curse} due to their lack of human-like inductive biases towards hierarchical structures \citep{mccoy-etal-2020-syntax}, phenomena also shared in more recent, larger LLMs \citep{tang-etal-2023-large, du2023shortcut, rizwan2026persistenteffectslexicalitylarge}.

\paragraph{Geometric analyses.}
To analyze whether the degradation stems from representation collapse, we compare frozen, mean-pooled last-layer states of each base and +PR model, reporting anisotropy and the number of principal components needed for 90\% of the variance in input embedding (PCA 90\%). We also train linear heads on the frozen representations (3 seeds).
We consider T5 on CoNLL 2003 and WNLI, BERT on RTE, and GPT-2 on CoNLL 2003.

\begin{figure}
    \centering
    \includegraphics[width=0.5\linewidth]{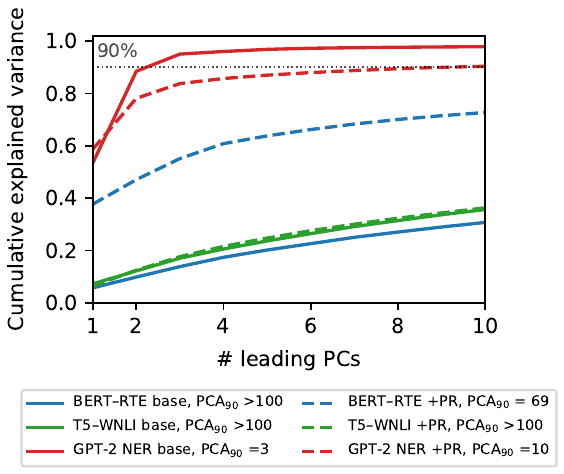}
    \caption{Cumulative variance explained by the first 10 principal components (solid = base, dashed = +PR; same sentences per pair). The higher the curve, the more collapsed the representation, and the lower the explained variance, the more high-dimensional and ``spread-out'' it is. Here, PR does not affect the dimensionality of T5, while it collapses BERT's and widens that of GPT-2.}
    \label{fig:geometric-analysis}
\end{figure}
Our findings are outlined in Figure \ref{fig:geometric-analysis}.
For T5, PR leaves the space largely intact on both CoNLL 2003 (anisotropy $0.429 \rightarrow 0.407$) and WNLI ($0.574 \rightarrow 0.601$), with PCA 90\% above 100 components throughout; WNLI probes underperform the majority baseline with both base and +PR.
BERT representations on RTE does collapse (anisotropy $0.680 \rightarrow 0.912$, PCA 90\% $>100 \rightarrow 69$), yet its probe performance changes little with +PR ($60.8$ $\rightarrow$ $59.8$, majority $52.7$).
GPT-2 shows the reverse: PR makes its highly anisotropic space slightly more isotropic ($0.985 \rightarrow 0.946$, PCA 90\% $3 \rightarrow 10$), but a NER head on base outperforms that on +PR by 9.6 F1 ($68.6 \rightarrow 59.0$), a gap that full fine-tuning largely closes.
We detail the setup of our analysis in Appendix \ref{sec:appendix-geometry}.

Overall, we find that PR doesn't affect T5's geometries, collapses BERT's, and widens GPT-2's.
At the same time, however, it appears that global geometry and downstream performance are not always correlated, and any decay or improvement in performance can be attributed to changes in task-relevant representations rather than to overall collapse in representation dimension.

\section{Conclusion}
We interpret our results to indicate that PR is an insufficient proxy for improving structural understanding. While previous work \citep{min-etal-2025-punctuation} found that PR improved downstream structure-related tasks for a single architecture using proprietary data, our results suggest that these benefits do not reliably generalize across architectures and tasks, and such success may be attributable to training- or data-specific features, rather than improved boundary prediction via PR.

LMs can be optimised to excel at boundary prediction (via PR), but in our experiments this improvement does not reliably carry over to downstream tasks that depend on linguistic structure. Changes in global representation geometry also do not consistently track these downstream effects.
We find that while PR remains accessible and scalable, relying on it to enrich language models' linguistic structure understanding appears largely ineffective.
  

\section*{Limitations}
\paragraph{Scale of language models studied.}
Our results only cover English monolingual language models with less than 1B parameters, substantially smaller than today's commercial and open-weight LLMs that process linguistic input for commercial and academic purposes.
While our findings are negative, the scope of the language models studied in our paper is nonetheless a limitation.
While language agnostic boundary predictor have been proposed \citep[e.g.][]{frohmann-etal-2024-segment}, we do not present our findings as a general claim on language or transformer-based language models as a whole, but rather that limited to English.

\paragraph{Operationalisation of structural representation quality as fine-tuned downstream task performance.}
A second limitation is our operationalisation of structural representation as downstream task performance.
Just as high representation quality may not directly translate to high downstream performance due to noise and dataset quality, high downstream performance does not necessarily indicate high representation quality.
Neural systems may optimise on spurious correlations to achieve high dataset scores despite their lack of task proficiency \citep{gururangan-etal-2018-annotation, mccoy-etal-2019-right, niven-kao-2019-probing}, and a ``lucky'' classifier initialisation may also be a factor that results in a misleadingly high task performance \citep{zhou-etal-2020-curse, mccoy-etal-2020-berts}.

\section*{Ethical Considerations}
We report our use of AI agents as coding assistants in our experimental implementation and as a tool to generate tables.
While we do not explicitly check for harmful content, our study only involves publicly available datasets and model weights.

\section*{Acknowledgments}
This work was made possible in part by support from the Fields Institute for Research in Mathematical Sciences and the Digital Research Alliance of Canada / Alliance de recherche numérique du Canada.


\small
\bibliographystyle{unsrtnat}
\bibliography{custom}









\newpage
\appendix
\section{Details on Base and +PR Performance}
\label{sec:appendix-results}

Table \ref{tab:downstream_and_glue_results} in Section \ref{sec:results-discussion} reports average downstream task performance over three runs with random classifier initialisation.
Here, we provide the result of each recorded run in Table \ref{tab:each-run-results} (structure-related tasks) and Table \ref{tab:each-run-results-glue} (GLUE).

\begin{table*}[t]
    \centering
    \small
    \begin{tabular}{l c cc cc cc}
\toprule
\textbf{Task (Dataset)} & \textbf{Run} & \multicolumn{2}{c}{\textbf{T5-base}} & \multicolumn{2}{c}{\textbf{GPT-2-medium}} & \multicolumn{2}{c}{\textbf{BERT-base}} \\
\cmidrule(lr){3-4} \cmidrule(lr){5-6} \cmidrule(lr){7-8}
 & & \textbf{base} & \textbf{+PR} & \textbf{base} & \textbf{+PR} & \textbf{base} & \textbf{+PR} \\
\midrule
\multirow{3}{*}{RE (TACRED)} & 1 & 62.52 & 60.46 & 64.34 & 65.19 & 64.90 & 64.23 \\
 & 2 & 63.63 & 59.33 & 64.59 & 65.41 & 67.37 & 64.65 \\
 & 3 & 63.26 & 61.62 & 64.04 & 66.21 & 65.70 & 64.59 \\
\addlinespace
\multirow{3}{*}{NER (CoNLL 2003)} & 1 & 89.39 & 89.07 & 83.43 & 78.55 & 90.51 & 90.92 \\
 & 2 & 89.86 & 89.25 & 82.09 & 80.94 & 90.80 & 91.05 \\
 & 3 & 89.09 & 89.09 & 83.45 & 76.66 & 91.25 & 91.15 \\
\addlinespace
\multirow{3}{*}{NER (GENIA)} & 1 & 77.22 & 76.43 & 48.70 & 48.32 & 76.42 & 75.51 \\
 & 2 & 77.86 & 77.71 & 49.23 & 49.71 & 76.30 & 75.25 \\
 & 3 & 77.99 & 77.74 & 47.06 & 49.28 & 76.43 & 75.02 \\
\addlinespace
\multirow{3}{*}{NER (OntoNotes5)} & 1 & 90.56 & 90.09 & 78.87 & 78.48 & 86.63 & 86.93 \\
 & 2 & 90.20 & 90.09 & 78.93 & 78.09 & 87.20 & 86.85 \\
 & 3 & 90.64 & 89.99 & 79.32 & 78.25 & 86.94 & 86.55 \\
\addlinespace
\multirow{3}{*}{POS (CoNLL 2000)} & 1 & 97.82 & 97.43 & 96.56 & 96.22 & 98.08 & 98.08 \\
 & 2 & 97.09 & 97.67 & 96.47 & 96.38 & 98.14 & 98.13 \\
 & 3 & 98.17 & 97.93 & 96.24 & 96.29 & 98.09 & 98.20 \\
\addlinespace
\multirow{3}{*}{SRL (CoNLL 2012)} & 1 & 87.57 & 87.04 & 82.26 & 82.30 & 88.34 & 87.91 \\
 & 2 & 87.42 & 87.27 & 82.28 & 82.07 & 88.24 & 87.98 \\
 & 3 & 87.47 & 87.08 & 82.25 & 82.12 & 88.20 & 87.95 \\
\addlinespace
\multirow{3}{*}{OpenIE (OIE2016)} & 1 & 66.95 & 66.00 & 42.93 & 44.89 & 83.82 & 82.20 \\
 & 2 & 66.54 & 63.62 & 48.46 & 44.32 & 85.09 & 83.23 \\
 & 3 & 66.76 & 65.08 & 50.58 & 46.15 & 83.56 & 82.89 \\
\bottomrule

\end{tabular}

\caption{Results of each run for the structure-related tasks, for the off-the-shelf models (base) and models after extended pretraining on PR (+PR). Scores are F1 using each dataset's default splits; each run uses a different random classifier initialisation. The means of these runs are reported in Table \ref{tab:downstream_and_glue_results}.
}
\label{tab:each-run-results}
\end{table*}

\begin{table*}[t]
    \centering
    \small
    \begin{tabular}{l c cc cc cc}
\toprule
\textbf{Task} & \textbf{Run} & \multicolumn{2}{c}{\textbf{T5-base}} & \multicolumn{2}{c}{\textbf{GPT-2-medium}} & \multicolumn{2}{c}{\textbf{BERT-base}} \\
\cmidrule(lr){3-4} \cmidrule(lr){5-6} \cmidrule(lr){7-8}
 & & \textbf{base} & \textbf{+PR} & \textbf{base} & \textbf{+PR} & \textbf{base} & \textbf{+PR} \\
\midrule
\multirow{3}{*}{SST-2} & 1 & 93.12 & 92.20 & 93.92 & 92.09 & 91.74 & 90.83 \\
 & 2 & 92.89 & 91.97 & 94.15 & 93.12 & 92.55 & 90.02 \\
 & 3 & 93.46 & 91.74 & 93.69 & 92.78 & 93.35 & 90.94 \\
\addlinespace
\multirow{3}{*}{QQP} & 1 & 90.40 & 89.08 & 90.33 & 90.25 & 91.14 & 90.71 \\
 & 2 & 90.36 & 89.29 & 90.35 & 90.26 & 91.14 & 90.65 \\
 & 3 & 90.32 & 89.21 & 90.33 & 90.13 & 91.09 & 90.86 \\
\addlinespace
\multirow{3}{*}{MNLI-m} & 1 & 85.86 & 83.62 & 85.02 & 84.00 & 83.98 & 83.52 \\
 & 2 & 85.95 & 83.66 & 85.45 & 83.79 & 84.48 & 83.44 \\
 & 3 & 85.82 & 83.49 & 85.33 & 84.10 & 83.73 & 83.04 \\
\addlinespace
\multirow{3}{*}{MNLI-mm} & 1 & 86.21 & 84.14 & 85.34 & 84.73 & 84.40 & 84.04 \\
 & 2 & 86.16 & 84.60 & 85.50 & 84.35 & 85.13 & 83.25 \\
 & 3 & 86.13 & 84.32 & 85.43 & 84.44 & 84.36 & 83.58 \\
\addlinespace
\multirow{3}{*}{QNLI} & 1 & 90.74 & 88.67 & 90.68 & 89.71 & 91.42 & 89.35 \\
 & 2 & 90.61 & 88.41 & 89.82 & 89.60 & 90.02 & 89.69 \\
 & 3 & 90.65 & 88.39 & 90.33 & 89.55 & 91.76 & 89.71 \\
\addlinespace
\multirow{3}{*}{RTE} & 1 & 54.51 & 58.48 & 66.43 & 60.29 & 63.90 & 59.57 \\
 & 2 & 61.01 & 62.82 & 65.70 & 60.29 & 66.06 & 61.37 \\
 & 3 & 51.26 & 53.79 & 67.51 & 57.76 & 63.90 & 62.09 \\
\addlinespace
\multirow{3}{*}{WNLI} & 1 & 49.30 & 50.70 & 35.21 & 23.94 & 33.80 & 40.85 \\
 & 2 & 52.11 & 54.93 & 36.62 & 33.80 & 43.66 & 47.89 \\
 & 3 & 47.89 & 50.70 & 39.44 & 29.58 & 23.94 & 33.80 \\
\addlinespace
\multirow{3}{*}{CoLA} & 1 & 44.13 & 37.73 & 49.38 & 42.27 & 59.81 & 51.80 \\
 & 2 & 44.40 & 37.37 & 49.38 & 47.25 & 58.05 & 52.64 \\
 & 3 & 43.83 & 32.91 & 50.21 & 41.26 & 56.78 & 51.03 \\
\addlinespace
\multirow{3}{*}{MRPC} & 1 & 70.59 & 72.79 & 79.17 & 78.68 & 85.05 & 77.21 \\
 & 2 & 70.83 & 73.77 & 79.17 & 78.68 & 86.03 & 79.90 \\
 & 3 & 71.08 & 74.51 & 78.68 & 76.23 & 87.50 & 81.86 \\
\addlinespace
\multirow{3}{*}{STS-B} & 1 & 76.18 & 77.95 & 85.40 & 82.68 & 89.16 & 86.91 \\
 & 2 & 73.71 & 78.74 & 84.33 & 83.24 & 89.50 & 87.51 \\
 & 3 & 70.11 & 79.00 & 85.81 & 82.30 & 88.60 & 86.93 \\
\bottomrule
\end{tabular}
    
\caption{Results of each run for the GLUE tasks, for the off-the-shelf models (base) and models after extended pretraining on PR (+PR). Scores are accuracy, except for CoLA (Matthews correlation) and STS-B (Pearson correlation), on the GLUE validation split; each run uses a different random classifier initialisation. The means of these runs are reported in Table \ref{tab:downstream_and_glue_results}.}
    \label{tab:each-run-results-glue}
\end{table*}

\section{Details on the FLOPs and Data Control}
\label{sec:appendix-flops-baseline}

In Section \ref{sec:results-discussion}, we report that additional training T5 with the language modeling objective degrades downstream task performance.
We provide the results from the control models in Table \ref{tab:flops-control}.

\section{Details on Geometric Analysis}
\label{sec:appendix-geometry}

In this Section, we detail our geometric analysis described in Section \ref{sec:results-discussion}.

\paragraph{Anisotropy.} We report anisotropy as the mean pairwise cosine similarity over all pairs of frozen sentence states in the sample.
An anisotropic value of 1 means every vector is equivalent, meaning that such representation is fully collapsed; 0 means the vectors are spread evenly in all directions, and the representation is isotropic.

\paragraph{PCA 90\%.} We run PCA once per model on the frozen states with at most 100 components and take the cumulative explained variance.
PCA 90\% is the smallest $k$ whose cumulative variance reaches $0.90$;
a value of $>100$ means the 90\% threshold was never reached within the 100 computed components.

\paragraph{Probe on frozen hidden states.} 
For our probe analysis, we mean-pool the last-layer hidden states (maximum length 128) over the same inputs for each base - +PR pair:
the hidden states are frozen and only the linear head weights are updated during the 3-seed (42, 123, and 456), 10-epoch training runs. 

\begin{table*}[t]
    \centering
    \small
    
    \begin{tabular}{l ccc}
    \toprule
    \textbf{Task (Dataset)} & \textbf{T5-base} & \textbf{+LM (1 epoch)} & \textbf{+LM (10 epochs)} \\
    \midrule
    SRL (CoNLL 2012) & 87.49 & 80.54 & 78.51 \\
    NER (CoNLL 2003) & 89.45 & 89.93 & 87.00 \\
    NER (GENIA) & 77.69 & 66.84 & 48.44 \\
    NER (OntoNotes5) & 90.47 & 86.65 & 82.69 \\
    OpenIE (OIE2016) & 66.75 & 33.99 & 16.49 \\
    \bottomrule
    \end{tabular}
    
\caption{Downstream performance (F1) of off-the-shelf T5-base (mean of three runs, as in Table \ref{tab:downstream_and_glue_results}) and of T5-base after continued training on the financial news corpus with the language modeling objective (+LM) for 1 and 10 epochs (single run).
}
\label{tab:flops-control}
\end{table*}

\end{document}